\documentclass[runningheads]{llncs}
\usepackage{graphicx}
\usepackage{wrapfig}
\newcommand{\xu}[1]{\textcolor{black}{#1}}

\newcommand{\red}[1]{\textcolor{black}{#1}}
\usepackage{tikz}
\usepackage{comment}
\usepackage{amsmath,amssymb} % define this before the line numbering.
\usepackage{color}
\usepackage{hyperref}
\usepackage{bm}
\usepackage{multicol,multirow}
\usepackage{algorithm,algorithmic}
\usepackage[accsupp]{axessibility}  % Improves PDF readability for those with disabilities.
\makeatletter
  \newcommand\tabcaption{\def\@captype{table}\caption}
\makeatother
\usepackage{subfigure}
\usepackage{booktabs}
\usepackage[marginal]{footmisc}

\begin{document}
\pagestyle{headings}
\mainmatter
\def\ECCVSubNumber{415}  % Insert your submission number here

\title{Recurrent Bilinear Optimization for Binary Neural Networks} 
\authorrunning{Xu et al.} 
\titlerunning{RBONN}
% INITIAL SUBMISSION 
%\begin{comment}
\author{Sheng~Xu\textsuperscript{1}$^{\dagger}$, Yanjing~Li\textsuperscript{1}$^{\dagger}$, Tiancheng~Wang\textsuperscript{1}, Teli~Ma\textsuperscript{2}, Baochang~Zhang\textsuperscript{1,3}$^{*}$, Peng~Gao\textsuperscript{2}, Yu~Qiao\textsuperscript{2}, Jinhu~L\"u\textsuperscript{1,3}, Guodong~Guo\textsuperscript{4,5}}
\institute{\textsuperscript{1} Beihang University, Beijing, China\\
\textsuperscript{2} Shanghai Artificial Intelligence Laboratory, Shanghai, China\\
\textsuperscript{3} Zhongguancun Laboratory, Beijing, China\\
\textsuperscript{4} Institute of Deep Learning, Baidu Research, Beijing, China\\
\textsuperscript{5} National Engineering Laboratory for Deep Learning Technology and Application, Beijing, China\\
\email{\{shengxu, yanjingli, bczhang\}@buaa.edu.cn}
}

%******************
\maketitle
\footnotetext{$\dagger$ Equal contribution.}
\footnotetext{$*$ Corresponding author.}

\begin{abstract}
Binary Neural Networks (BNNs) show great promise for real-world embedded devices. As one of the critical steps to achieve a powerful BNN, the scale factor calculation plays an essential role in reducing the performance gap to their real-valued counterparts. However, existing BNNs neglect the intrinsic bilinear relationship of real-valued weights and scale factors, resulting in a sub-optimal model caused by an insufficient training process. To address this issue, {\bf {\em Recurrent Bilinear Optimization}} is proposed to improve the learning process of {\bf {\em BNNs}} (RBONNs) by associating the intrinsic bilinear variables in the back propagation process. Our work is the first attempt to optimize BNNs from the bilinear perspective. Specifically, we employ a recurrent optimization and Density-ReLU to sequentially backtrack the sparse real-valued weight filters, which will be sufficiently trained and reach their performance limits based on a controllable learning process. We obtain robust RBONNs, which show impressive performance over state-of-the-art BNNs on various models and datasets. Particularly, on the task of object detection, RBONNs have great generalization performance.
Our code is open-sourced on \url{https://github.com/SteveTsui/RBONN}.

\end{abstract}

\section{Introduction}
Computer vision has been rapidly promoted, with the widespread application of convolutional neural networks (CNNs) in image classification \cite{imagenet15,gao2022convmae}, semantic segmentation \cite{ssegmetation}, and object detection \cite{voc2007,coco2014}. It does, however, come with a huge demand for memory and computing resources. These computation and memory costs are incompatible with the computing capabilities of devices, particularly those with low resources, {\em  e.g.}, mobile phones and embedded devices. As a result, substantial research has been invested to reduce storage and computation cost. To accomplish this, a number of compression methods for efficient inference have been proposed, including network pruning \cite{lecun1990optimal,he2018soft,li2016pruning}, low-rank decomposition \cite{denil2013predicting,lin2017espace}, network quantization \cite{rastegari2016xnor,liu2018bi,lin2021siman}, and knowledge distillation \cite{romero2014fitnets}. Network quantization, for example, is particularly well suited for using on embedded devices since it decreases the bit-width of network weights and activations. Binarization, a particularly aggressive kind of quantization, reduces CNN parameters and activations into 1 bit, reducing memory usage by $32 \times$ and calculation costs by $58 \times$ \cite{rastegari2016xnor}. 
Binarized neural networks (BNNs) are employed for a wide range of applications, such as image classification \cite{rastegari2016xnor,liu2018bi,liu2020reactnet}, object detection \cite{wang2020bidet,xu2021layer} and point cloud recognition \cite{xu2021poem}. 
With high energy-efficiency, they are potent to be directly applied on AI chips.
However, due to the limited representation capabilities, BNNs' performance remains considerably inferior to that of their real-valued counterparts.

\begin{figure*}[t]
\centering
\includegraphics[scale=.3]{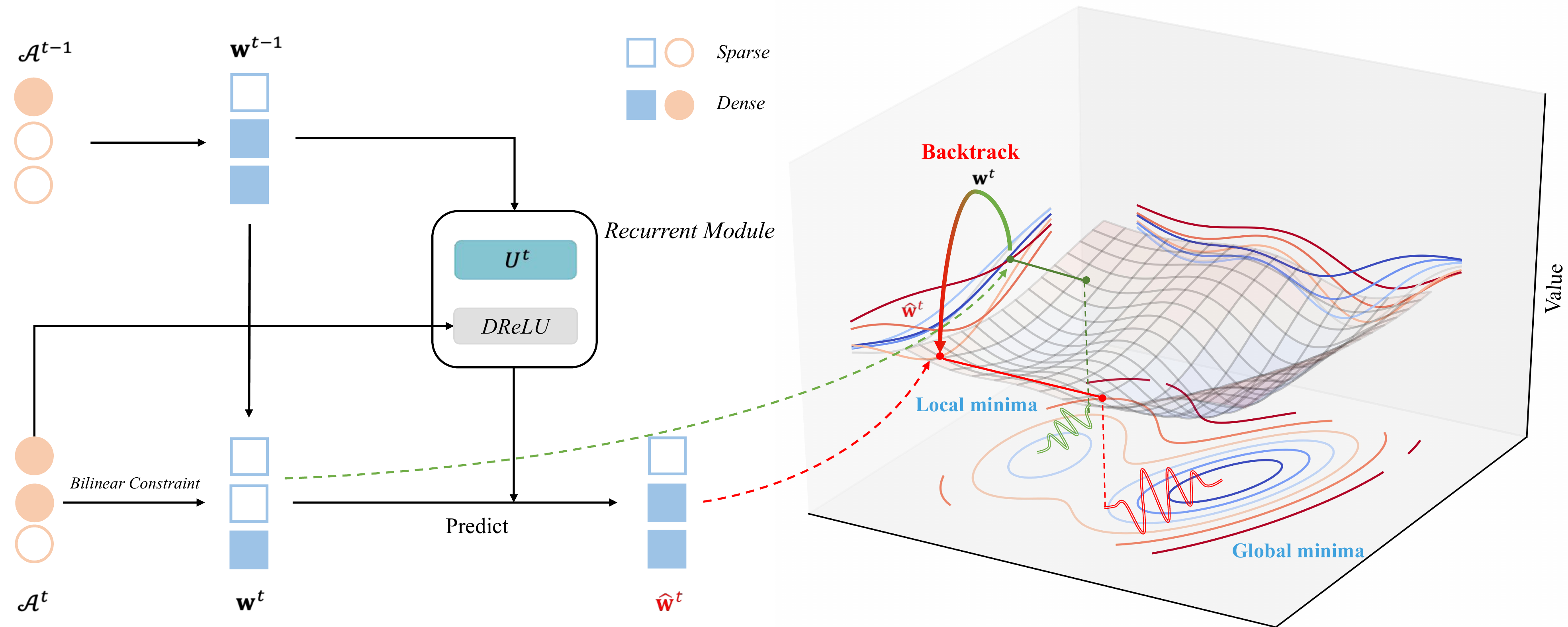}
\caption{An illustration of the RBONN framework. Conventional gradient-based algorithms assume that the hidden variables in bilinear models are independent, which  causes an insufficient training of ${\bf w}$ due to neglecting the relationship with  $\mathcal{\bm{A}}$ as shown in the loss surface (right part). Our RBONN can help ${\bf w}$ escape from local minima (green dotted line) and achieve a better solution (red dotted line).
}
\label{framework}
\end{figure*}

Previous methods \cite{gu2019projection,liu2019rbcn} compute scale factors by approximating the real-valued weight filter ${\bf w}$ such that ${\bf w} \approx \alpha \circ {{\bf b}}^{{\bf w}}$, where $\alpha\in \mathbb{R}_+$ is the scale factor (vector) and ${{\bf b}}^{{\bf w}}={\rm sign}({\bf w})$ to enhance the representation capability of BNNs. In essence, the approximation can be considered as a bilinear optimization problem with the objective function as
\begin{equation}
\mathop {\arg \min }\limits_{\mathbf{{\bf w}, {\alpha}}} G({\bf w}, \alpha) = \|{\bf w}-\alpha\circ{{\bf b}}^{{\bf w}}\|^2_2+R({\bf w}),\nonumber
\end{equation}
or
\begin{equation} 
\mathop {\arg \min }\limits_{\mathbf{{\bf w}, \mathcal{\bm{A}}}} G({\bf w}, \mathcal{\bm{A}})=\|{{\bf b}}^{{\bf w}} - \mathcal{\bm{A}}{\bf w}\|^2_2+R({\bf w}),
\label{1}
\end{equation}
where $\mathcal{\bm{A}}=diag(\frac{1}{\alpha_1},\cdots, \frac{1}{\alpha_N})$, $N$ is the number of elements in $\alpha$. $\circ$ denotes the channel-wise multiplication, and $R(\cdot)$ represents the regularization, typically the $\ell_1$ or $\ell_2$ norm. {$G({\bf w},\mathcal{A})$ includes a bilinear form of $\mathcal{A}{\bf w}$ widely used in the field of computer vision \cite{del2011bilinear,Liu2017learning,Huang2017Data}. Note that the bilinear function is $\mathcal{A} {\bf w}$ rather than $G({\bf w},\mathcal{A})$ in Eq. \ref{1}.} Eq. \ref{1} is rational for BNNs with $\mathcal{\bm{A}}$ and ${\bf w}$ as bilinear coupled variables, since ${\bf w}$ is the variable and ${\bf b}^{\bf w}$ is just the sign of ${\bf w}$.
%However, the coupling of different variables (${\bf w}$ and $\mathcal{\bm{A}}$) is seldom investigated in the literature, so that the collaborative nature of different variables for optimization is neglected, leading to a sub-optimal solution to calculate BNNs. 
However, such bilinear constraints will lead to an asynchronous convergence problem and   directly affect the learning process of $\mathcal{A}$ and ${\bf w}$. We can know that the variable with a slower convergence speed (usually ${\bf w}$) is {\bf not as sufficiently} trained as another faster one. Moreover, BNNs are based on non-convex optimization, and will suffer more from local minima problem due to such an asynchronous convergence. A powerful instance is that ${\bf w}$ will tendentiously fall into the local optimum with low magnitude when the magnitude of $\mathcal{\bm{A}}$ is much larger than $0$ (due to ${\bf b}^{\bf w}\in\{-1, +1\}$). On the contrary, ${\bf w}$ will have a large magnitude and thus slowly converge when elements of $\mathcal{\bm{A}}$ are close to $0$. 

In this paper, we introduce a recurrent bilinear optimization for binary neural networks (RBONNs) by learning the coupled scale factor and real-valued weight end-to-end. More specifically, recurrent optimization can efficiently backtrack the weights, which will be more sufficiently trained than conventional methods. To this end, a Density-ReLU (DReLU) is introduced to activate the optimization process based on the density of the variable  $\mathcal{\bm{A}}$. In this way, we achieve a controlled learning process with a backtracking mechanism by considering the interaction of variables, thus avoiding the local minima and reaching the performance limit of BNNs, as shown in Fig. \ref{framework}. Our contributions can be summarized as 
\begin{itemize}
    \item \red{We are the first attempt to address BNNs as a bilinear optimization problem.} A recurrent bilinear optimization is introduced for BNNs (RBONNs), {which can more sufficiently train BNNs and approach its performance limit.}
    \item A Density-ReLU (DReLU) is introduced to activate the optimization process based on the interaction of BNN variables, which can efficiently improve the training process of BNNs.
   \item \red{Extensive experiments show that the proposed RBONN outperforms state-of-the-art BNNs on a variety of tasks, including image classification and object detection. For instance, on ImageNet, the 1-bit ResNet-18 achieved by RBONN obtains 66.7\% Top-1 accuracy, outperforming all prior BNNs and achieving a new state-of-the-art.}
\end{itemize}

\section{Related Work}
\noindent\textbf{Bilinear Models in Deep Learning. } Under certain circumstances, bilinear models can be used in CNNs. One important application, network pruning, is among the hottest topics in the deep learning community \cite{lin2019towards,Liu2017learning}. The vital feature maps and related channels are pruned using bilinear models \cite{Liu2017learning}. 
Iterative methods, {\em e.g.}, the Fast Iterative Shrinkage-Thresholding Algorithm (FISTA) \cite{lin2019towards} and the Accelerated Proximal Gradient (APG) \cite{Huang2017Data} can be used to prune bilinear-based networks.
Many deep learning applications, such as fine-grained categorization \cite{lin2015bilinear,li2017factorized}, visual question answering (VQA) \cite{yu2017multi}, and person re-identification \cite{suh2018part}, are promoted by embedding bilinear models into CNNs, which model pairwise feature interactions and fuse multiple features with attention.

\noindent\textbf{Binary Neural Network.}
Based on BinaryConnect, BinaryNet \cite{courbariaux2016binarized} trains CNNs with binary parameters. 
By binarizing the weights and inputs of the convolution layer, the XNOR-Net \cite{rastegari2016xnor} improves the efficiency of CNNs.
Based on a discrete backpropagation process, a binarization approach is proposed in \cite{gu2019projection} to learn improved BNNs.

ReActNet \cite{liu2020reactnet} substitutes the traditiona sign function and PReLU \cite{he2015delving} with RSign and RPReLU based on learnable thresholds, resulting in improved BNN performance. RBNN \cite{lin2020rotated} rotates the real-valued weights for fruitful information, thus improving the feature representation of BNNs. SLB \cite{yang2020searching} introduces the NAS \cite{liu2018darts} into the binarization of weights.

\xu{Unlike prior work, our work is the first attempt to solve BNNs as a bilinear optimization problem. \red{We achieve training BNNs sufficiently to bridge the performance gap between them and their real-valued equivalents.}}

\section{Methodology}
In this section, we describes RBONN in detail. We first describe the bilinear model of BNNs, then introduce a recurrent bilinear optimization method to calculate BNNs, followed by a summary of the whole training process. For a better presentation of our approach, we first briefly describe the preliminaries.

\subsection{Preliminaries}
In a specific convolution layer, ${\bf w}\in \mathbb{R}^{C_{out}\times C_{in}\times K \times K}$, ${\bf a}_{in}\in \mathbb{R}^{C_{in} \times W_{in} \times H_{in}}$, and ${\bf a}_{out}\in \mathbb{R}^{C_{out} \times W_{out} \times H_{out}}$ represent its weights and feature maps, where  $C_{in}$ and $C_{out}$ represents the number of channels. $(H, W)$ are the height and width of the feature maps, and $K$ denotes the kernel size. We then have

\begin{equation}
{\bf a}_{out} = {\bf a}_{in} \otimes {\bf w},
\label{2}
\end{equation}
where $\otimes$ is the convolution operation. We omit the batch normalization (BN) and activation layers for simplicity. The 1-bit model aims to quantize $ {\bf w}$ and $ {\bf a}_{in}$ into ${{\bf b}^{{\bf w}}}\in \{-1,+1\}^{C_{out}\times C_{in}\times K \times K}$ and ${{\bf b}^{{\bf a}_{in}}}\in\{-1,+1\}^{C_{in} \times W_{in} \times H_{in}}$ using the efficient XNOR and Bit-count operations to replace real-valued operations. Following \cite{rastegari2016xnor}, the forward process of the BNN is
\begin{equation}
\small
{{\bf a}_{out}} = {\bf b}^{{\bf a}_{in}} \odot (\mathcal{\bm{A}}^{-1}{\bf b}^{{\bf w}}),
\label{3}
\end{equation}
where $\odot$ denotes the efficient XNOR and Bit-count operations. We divide the data flow in BNNs into units for detailed discussions. In BNNs, the original output ${\bf a}_{out}$ is first scaled by a channel-wise scale 
factor (matrix) $\mathcal{\bm{A}}=diag(\frac{1}{\alpha_1},\cdots,\\\frac{1}{\alpha_{C_{out}}})\in\mathbb{R}_+^{C_{out}\times C_{out}}$ to modulate the amplitude of full-precision 
counterparts. It then enters several non-linear layers, {\em e.g.}, BN layer, non-linear activation layer, and max-pooling layer. We omit these for simplification. And then, the output is  ${{\bf a}_{out}}$ via the sign function. Then, ${\bf b}^{{\bf a}_{out}}$ can be utilized for the efficient operations of the next layer.

\subsection{Bilinear Model of BNNs}
We formulate the optimization of BNNs as
\begin{equation}
\mathop {\arg \min }\limits_{\mathbf{{\bf w}, \mathcal{\bm{A}}}} L_S({\bf w},\mathcal{\bm{A}})+\lambda G({\bf w}, \mathcal{\bm{A}}),
\label{5}
\end{equation}
where $\lambda$ is the hyper-parameter. $G$ contains the bilinear part as mentioned in Eq. \ref{1}. ${\bf w}$ and $\mathcal{\bm{A}}$ formulate a pair of coupled variables. Thus, the conventional gradient descent method can be used to solve the bilinear optimization problem as  
\begin{equation}
\mathcal{\bm{A}}^{t+1} = |\mathcal{\bm{A}}^{t} - \eta_1\frac{\partial L}{\partial \mathcal{\bm{A}}^{t}}|,
\label{6}
\end{equation}
\begin{equation}
\small
\begin{aligned}
(\frac{\partial L}{\partial \mathcal{\bm{A}}^{t}})^T&=(\frac{\partial L_S}{\partial \mathcal{\bm{A}}^{t}})^T + \lambda (\frac{\partial G}{\partial \mathcal{\bm{A}}^{t}})^T,\\
&=(\frac{\partial L_S}{\partial {\bf a}_{out}^{t}}\frac{\partial {\bf a}_{out}^{t}}{\partial \mathcal{\bm{A}}^{t}})^T + \lambda {{{\bf w}^{t}}}(\mathcal{\bm{A}}^{t}{\bf w}^{t}- {{\bf b}^{{\bf w}^{t}}})^T,\\
&=(\frac{\partial L_S}{\partial {\bf a}_{out}^{t}})^T({\bf b}^{{\bf a}^t_{in}} \odot {\bf b}^{{\bf w}^t})(\mathcal{\bm{A}}^t)^{-2} + \lambda{\bf w^t}\hat{G}({\bf w}^t, \mathcal{\bm{A}}^t),
\end{aligned}
\label{7}
\end{equation}
where $\eta_1$ is the learning rate, $\hat{G}({\bf w}^t, \mathcal{\bm{A}}^t)=(\mathcal{\bm{A}}^{t}{\bf w}^{t}- {{\bf b}^{{\bf w}^{t}}})^T$. {Conventional gradient descent algorithm for bilinear models iteratively optimizes one variable while keeping the other fixed. This is actually a sub-optimal solution due to  ignoring the relationship of the two hidden variables in optimization. For example, when ${\bf w}$ approaches zero due to the sparsity regularization term $R({\bf w})$,  $\mathcal{\bm{A}}$ will have a  larger magnitude due to $G$ (Eq. \ref{1}). Consequently, both the first and second values of Eq. \ref{7} will be suppressed dramatically, causing  the gradient vanishing problem for $\mathcal{\bm{A}}$. Contrarily, if $\mathcal{\bm{A}}$  changes little during optimization,  ${\bf w}$ will also suffer from the vanished gradient problem due to the supervision of $G$, causing a local minima.} 
Due to the coupling relationship of ${\bf w}$ and $\mathcal{\bm{A}}$, the  gradient calculation for ${\bf w}$ is challenging.

\subsection{Recurrent Bilinear Optimization}
{We solve the problem in Eq. \ref{1} from a new perspective that ${\bf w}$ and $\mathcal{\bm{A}}$ are coupled. We aim to prevent  $\mathcal{\bm{A}}$ from going denser and ${\bf w}$ from going sparser, as analyzed above. Firstly, based on the chain rule and its notations in \cite{petersen2008matrix}, we have the scalar form of the update rule for $\widehat{{\rm w}}_{i,j}$ as}
\begin{equation}
\begin{aligned}
\widehat{{\rm w}}_{i,j}^{t+1} &= {\rm w}_{i,j}^{t} - \eta_2\frac{\partial L_S}{\partial {\rm w}_{i,j}^{t}} - \eta_2\lambda (\frac{\partial G}{\partial {\rm w}_{i,j}^{t}} + Tr((\frac{\partial G}{\partial  \mathcal{\bm{A}}^t})^T\frac{\partial \mathcal{\bm{A}}^{t}}{\partial {\rm w}_{i,j}^{t}})),\\
&={\rm w}_{i,j}^{t+1} - \eta_2\lambda Tr({\bf w}^{t}\hat{G}({\bf w}^t, \mathcal{\bm{A}}^t)\frac{\partial \mathcal{\bm{A}}^{t}}{\partial {\rm w}^{t}_{i,j}}),
\end{aligned}
\label{9}
\end{equation}
which is based on ${\rm w}_{i,j}^{t+1} = {\rm w}_{i,j}^{t} - \eta_2\frac{\partial L}{\partial {\rm w}_{i,j}^{t}}$. {$\hat{{\bf w}}^{t+1}$ denotes ${\bf w}$ at the $t+1$-th iteration when considering the coupling of ${\bf w}$ and $\mathcal{A}$. When computing the gradient of the coupled variable ${\bf w}$, the gradient of its coupled variable $\mathcal{A}$ should also be considered using the chain rule. Vanilla ${{\bf w}}^{t+1}$ denotes the computed ${\bf w}$ at $t+1$-th iteration without considering the coupling relationship.} Here we denote $I=C_{out}$ and $J = C_{in} \times K \times K$ for simplicity. With writing ${\bf w}$ into a row vector $[{\bf w}_1, \cdots ,{\bf w}_I ]^T$ and writing $\hat{G}$ into a column vector $[\hat{g}_1, \cdots, \hat{g}_I]$ and using $i = 1, \cdots, I$ and $j = 1, \cdots, J$, we can see that $\mathcal{A}_{i,i}$ and ${\rm w}_{nj}$ are independent when ${\forall}n\ne j$. Omitting superscript $\cdot^t$, we have the $i$-th component of $\frac{\partial \mathcal{A}}{\partial {\bf w}}$ as 
	\begin{equation}
	\small
		(\frac{\partial \mathcal{A}}{\partial {\bf w}})_i=
		\begin{bmatrix} 
		0 &...&.&...& 0 \\
		.&&.&&.\\
		\frac{\partial \mathcal{A}_{i,i}}{\partial {\rm w}_{i,1}} &...&\frac{\partial \mathcal{A}_{i,i}}{\partial {\rm w}_{i,j}}&...& \frac{\partial \mathcal{A}_{i,i}}{\partial {\rm w}_{i,J}}\\
		.&&.&&.\\ 
		0 &...&.&...& 0
		\end{bmatrix}, 
		\label{eq:8}
	\end{equation} 
	we can derive
	\begin{equation}
	\small
		{\bf w}\hat{G}({\bf w, \mathcal{A}})=
		\begin{bmatrix} 
		{\bf w}_1\hat{g}_1 &...&{\bf w}_1\hat{g}_i&...& {\bf w}_1\hat{g}_I \\
		.&&.&&.\\
		.&&.&&.\\
		.&&.&&.\\ 
		{\bf w}_I\hat{g}_1 &...&{\bf w}_I\hat{g}_i&...& {\bf w}_I\hat{g}_I
		\end{bmatrix}.
		\label{eq:9}
	\end{equation}
Combine Eq.~\ref{eq:8} and Eq.~\ref{eq:9}, we get
	\begin{equation}
	\small
		{\bf w}\hat{G}({\bf w, \mathcal{A}})(\frac{\partial \mathcal{A}}{\partial {\bf w}})_i=
		\begin{bmatrix} 
		{\bf w}_1\hat{g}_i\frac{\partial \mathcal{A}_{i,i}}{\partial {\rm w}_{i,1}}&...&.&...&{\bf w}_1\hat{g}_i\frac{\partial \mathcal{A}_{i,i}}{\partial {\rm w}_{i,j}} \\
		.&&.&&.\\
		{\bf w}_i\hat{g}_i\frac{\partial \mathcal{A}_{i,i}}{\partial {\rm w}_{i,1}}&...&.&...&{\bf w}_i\hat{g}_i\frac{\partial \mathcal{A}_{i,i}}{\partial {\rm w}_{i,J}}\\
		.&&.&&.\\ 
		{\bf w}_I\hat{g}_i\frac{\partial \mathcal{A}_{i,i}}{\partial {\rm w}_{i,1}}&...&.&...&{\bf w}_I\hat{g}_i\frac{\partial \mathcal{A}_{i,i}}{\partial {\rm w}_{iJ}}
		\end{bmatrix}.
		\label{eq:10}
	\end{equation}
After that, the $i$-th component of the trace item in Eq.~\ref{9} is then calculated by:

	\begin{equation}
	\small
		Tr[{\bf w} \hat{G} (\frac{\partial \mathcal{A}}{\partial {\bf w}})_i]={\bf w}_i\hat{g}_i\sum_{j=1}^J\frac{\partial \mathcal{A}_{i,i}}{\partial {\rm w}_{i,j}}
		\label{eq:11}
	\end{equation}
	Combining Eq.~\ref{9} and  Eq.~\ref{eq:11}, we can get
	\begin{equation}
	\small
		\label{eq:12}
		\begin{aligned}
		\hat{{\bf w}}^{t+1}&={\bf w}^{t+1}-\eta_2 \lambda 
		\begin{bmatrix} 
		\hat{g}_1^t\sum_{j=1}^J\frac{\partial \mathcal{A}_{1,1}^t}{\partial {\rm w}_{1,j}^t}\\
		.\\
		.\\
		.\\ 
		\hat{g}_I^t\sum_{j=1}^J\frac{\partial \mathcal{A}_{I,I}^t}{\partial {\rm w}_{I,j}^t}
		\end{bmatrix}
		\circledast
		\begin{bmatrix} 
		{\bf w}_1^t\\
		.\\
		.\\
		.\\ 
		{\bf w}^t_I
		\end{bmatrix}\\    
		&={\bf w}^{t+1}+\eta_2 \lambda \bm{d}^t \circledast {\bf w}^t,	
		\end{aligned}
	\end{equation}
%where $i$ and $j$ are the coordination variables subject to $0\leq i \leq C_{in}\times K \times K$ and $0\leq j \leq C_{out}$. By introducing Eq. \ref{8} into Eq. \ref{9}, we have
%\begin{equation}
%\widehat{{\bf w}}^{t+1} ={\bf w}^{t+1} + %\eta_2\bm{d}^t\circledast{\bf w}^t
%\label{10}
%\end{equation}
where $\eta_2$ is the learning rate of real-valued weight filters ${\bf w}_i$, $\circledast$ denotes the Hadamard product. 
%Because {the magnitude of} ${\bf w}$ and $\mathcal{\bm{A}}$ share a {negative correlation},
We take $\bm{d}^t = -[\hat{g}^t_1\sum_{j=1}^J\frac{\partial \mathcal{\bm{A}}^t_{1,1}}{\partial {\rm w}^t_{1,j}},\cdots,\hat{g}^t_I\sum_{j=1}^J\frac{\partial \mathcal{\bm{A}}^t_{i,i}}{\partial {\rm w}^t_{I,j}}]^T$, {which is unsolvable and undefined in the back propagation of BNNs. To address this issue, we employ a recurrent model to approximate $d^t$ and have}
%{Conventional gradient under bilinear relationship will constrains ${\bf w}$ to gain its sparsity during training. To overcome this obstacle, we introduce a recurrent mechanism into the training of ${\bf w} $ as}
\begin{equation}
\hat{{\bf w}}^{t+1} = {\bf w}^{t+1} +
U^t\circ DReLU({\bf w}^t, \mathcal{\bm{A}}^{t}),
\label{11}
\end{equation}
and
\begin{equation}
    {\bf w}^{t+1} \leftarrow \hat{{\bf w}}^{t+1},
\label{11*}
\end{equation}
where we introduce a hidden layer with channel-wise learnable weights $U\in \mathbb{R}_+^{C_{out}}$ to recurrently backtrack the ${\bf w}$. To realize a controllable recurrent optimization, we present $DReLU$ to supervise such an optimization process. We channel-wise implement $DReLU$ as 
\begin{equation}
\label{12}
DReLU({\bf w}_i, \mathcal{\bm{A}}_i)=
\begin{cases}
{\bf w}_i & if\ (\neg D({\bf w}'_i))\wedge D(\mathcal{\bm{A}}_i)=1,\\
0& otherwise,
\end{cases}
\end{equation}
where ${\bf w}' = diag(\|{\bf w}_1\|_1,\cdots,\|{\bf w}_{C_{out}}\|_1)$. And we judge when an asynchronous convergence happens in the optimization based on $(\neg D({\bf w}'_i))\wedge D(\mathcal{\bm{A}}_i)=1$, where the density function is defined as 
\begin{equation}
\label{13}
D(\bm{x}_i)=
\begin{cases}
1 & if\;\;ranking(\sigma(\bm{x})_i)\textgreater \mathcal{T},\\
0& otherwise,
\end{cases}
\end{equation}
where $\mathcal{T}$ is defined by $\mathcal{T} = int(C_{out}\times\tau)$. $\tau$ is the hyper-parameter denoting the threshold. $\sigma(\bm{x})_i$ denotes the $i$-th eigenvalue of diagonal matrix $\bm{x}$, and $\bm{x}_i$ denotes the $i$-th row of matrix $\bm{x}$. Finally, we define the optimization of $U$ and as
\begin{equation}
{U}^{t+1} = |U^{t} - \eta_3 \frac{\partial L}{\partial {U}^{t}}|,
\label{14}
\end{equation}
\begin{equation}
\label{15}
\frac{\partial L}{\partial {{U}^{t}}} =  \frac{\partial L_S}{\partial {{\bf w}^{t}}}\circ DReLU({\bf w}^{t-1}, \mathcal{\bm{A}}^t),
\end{equation}
where $\eta_3$ is the learning rate of $U$. We elaborate on the training process of RBONN outlined in Algorithm \ref{algo}.

\begin{algorithm}[t]
\caption{RBONN training.}
{\bf Input:} 
a minibatch of inputs and their labels, real-valued weights ${\bf w}$, recurrent model weights $U$, scale factor matrix $\mathcal{\bm{A}}$, learning rates $\eta_1$, $\eta_2$ and $\eta_3$.\\
{\bf Output:}
updated real-valued weights ${\bf w}^{t+1}$, updated scale factor matrix $\mathcal{\bm{A}}^{t+1}$, and updated recurrent model weights $U^{t+1}$.
\begin{algorithmic}[1]
	\WHILE{Forward propagation}
	\STATE ${\bf b}^{{\bf w}^t} \gets$ ${\rm sign}({\bf w}^t)$.
	\STATE ${\bf b}^{{\bf a}_{in}^t} \gets$ ${\rm sign}({\bf a}_{in}^t)$.
	\STATE Features calculation using Eq.~\ref{3}
    \STATE Loss calculation using Eq.~\ref{5}
	\ENDWHILE
	\WHILE{Backward propagation}
	    \STATE Computing $\frac{\partial L}{\partial \mathcal{\bm{A}}^{t}}$, $\frac{\partial L}{\partial {\bf w}^{t}}$, and $\frac{\partial L}{\partial U^{t}}$ using Eq. \ref{7} , \ref{9} and \ref{15}.
        \STATE Update $\mathcal{\bm{A}}^{t+1}$, ${\bf w}^{t+1}$, and $U^{t+1}$ according to Eqs. \ref{6}, \ref{11}, and \ref{14}, respectively.
	\ENDWHILE
\end{algorithmic}
\label{algo}
\end{algorithm}

\subsection{{Discussion}}
{In this section, we first review the related methods on ``gradient approximation" of BNNs, and then further discuss the difference of RBONN with the related methods and analyze the effectiveness of the proposed RBONN.}

{In particular, BNN \cite{courbariaux2016binarized} directly unitize the Straight-Through-Estimator in training stage to calculate the gradient of weights and activations as }
\begin{equation}
\begin{aligned}
    \frac{\partial {\bf b}^{{\bf w}_{i,j}}}{\partial {\bf w}_{i,j}} = {1}_{|{\bf w}_{i,j}|<{1}}, \frac{\partial {\bf b}^{{\bf a}_{i,j}}}{\partial {\bf a}_{i,j}} = {1}_{|{\bf a}_{i,j}|<{1}}
\end{aligned}
\end{equation}
{which suffers from an obvious gradient mismatch between
the gradient of the binarization function. Intuitively, Bi-Real Net \cite{liu2018bi} designs an approximate binarization function can help to relieve the gradient mismatch in the backward propagation as}
\begin{equation}
    \frac{\partial {\bf b}^{{\bf a}_{i,j}}}{\partial {\bf a}_{i,j}}=\left\{ \begin{matrix}
      \textcolor[rgb]{1,1,1}{1.} 2+2 {\bf a}_{i,j},\qquad-1\le {\bf a}_{i,j}<0,  \\
       2-2 {\bf a}_{i,j},\quad\quad0\le {\bf a}_{i,j}<1,  \\
       \begin{matrix}
      \textcolor[rgb]{1,1,1}{1} 0,\qquad\qquad\quad{{otherwise}}, & {}  \\
    \end{matrix}  \\
    \end{matrix} \right.
\end{equation}
{which is termed as ApproxSign function and used for back-propagation gradient calculation of the activation. Compared to the traditional STE, ApproxSign has a close shape to that of the original binarization function sign, and thus the activation gradient error can be controlled to some extent. Likewise, CBCN \cite{liu2019circulant} applies an approximate function to address the gradient mismatch from the sign function. MetaQuant \cite{chen2019metaquant} introduces Metalearning to learning the gradient error of weights by a neural network. The IR-Net \cite{qin2020forward} includes a self-adaptive Error Decay Estimator (EDE) to reduce the gradient error in training, which considers different requirements on different stages of the training process and balances the update ability of parameters and reduction of gradient error. RBNN \cite{lin2020rotated} proposes a training-aware approximation of the sign function for gradient backpropagation.}

{In summary, prior arts focus on approximating the gradient derived from $\frac{\partial {\bf b}^{\bf a}}{\partial {\bf a}_{i,j}}$ or $\frac{\partial {\bf b}^{\bf w}}{\partial {\bf w}_{i,j}}$. Differently, our approach focuses on a different perspective of gradient approximation, {\em i.e.}, gradient from $\frac{\partial G}{\partial {\bf w}_{i,j}}$. Our goal is to decouple $\mathcal{A}$ and ${\bf w}$ to improve the gradient calculation of ${\bf w}$. RBONN manipulates ${\bf w}$'s gradient from its bilinear coupling variable $\mathcal{A}$ ($\frac{\partial G(\mathcal{A})}{\partial {\bf w}_{i,j}}$). More specifically, our RBONN can be combined with prior arts, by comprehensively considering $\frac{\partial L_S}{\partial {\bf a}_{i,j}}$, $\frac{\partial L_S}{\partial {\bf w}_{i,j}}$ and $\frac{\partial G}{\partial {\bf w}_{i,j}}$ in the back propagation process. }

\section{Experiments}
\label{secexp}
Our RBONNs are evaluated first on image classification and then on object detection tasks. First, we introduce the implementation details of RBONNs. Then we validate the effectiveness of components in the ablation study. Finally, we illustrate the superiority of RBONNs by comparing our method with state-of-the-art BNNs on various tasks. 

\subsection{Datasets and Implementation Details}
\label{secdata}
\textbf{Datasets.}
For its huge scope and diversity, the ImageNet object classification dataset \cite{imagenet15} is more demanding, which has 1000 classes, 1.2 million training photos, and 50k validation images.

Natural images from 20 different classes are included in the VOC datasets. We use the VOC {\tt trainval2007} and VOC {\tt trainval2012} sets to train our model, which contains around 16k images, and the VOC {\tt test2007} set to evaluate our IDa-Det, which contains 4952 images. We utilize the mean average precision (mAP) as the evaluation matrices, as suggested by \cite{voc2007}.

The COCO dataset includes images from 80 different categories. 
All of our COCO dataset experiments are performed on the object detection track of the COCO {\tt trainval35k} training dataset, which consists of 80k images from the COCO {\tt train2014} dataset and 35k images sampled from the COCO {\tt val2014} dataset.  
We report the average precision (AP) for IoUs$\in [0.5:0.05:0.95]$, designated as mAP$@[.5,.95]$, using COCO's standard evaluation metric. For further analyzing our method, we also report AP$_{50}$, AP$_{75}$, AP$_s$, AP$_m$, and AP$_l$.

\noindent\textbf{Implementation Details.} 
\red{PyTorch \cite{paszke2017automatic} is used to implement RBONN. We run the experiments on 4 NVIDIA GTX 2080Ti GPUs with $11$ GB memory.} Following \cite{liu2018bi}, we retain the first layer, shortcut, and last layer in the networks as real-valued. We modify the architecture of the BNNs with extra shortcuts, and PReLU \cite{he2015delving} following \cite{liu2018bi} and \cite{gu2019projection}, respectively.

\xu{For the image classification task, ResNets \cite{he2016deep} and ReActNets \cite{liu2020reactnet} are employed as the backbone networks to build our RBONNs. We offer two implementation setups for fair comparison. \red{First, we use \textbf{one-stage training} on ResNets, using Adam as the optimization algorithm, and a weight decay of $1e\!\!-\!\!{5}$.} $\eta_1$ and $\eta_2$ are both set to $1e\!\!-\!\!{3}$. $\eta_3$ is set as $1e\!\!-\!\!{4}$. The learning rates are optimized by the annealing cosine learning rate schedule. The number of epochs is set as 200. Then, we employ \textbf{two-stage training} on ReActNets following \cite{liu2020reactnet}. Each stage counts 256 epochs. Thus the number of epochs is set as 512. In this implementation, Adam is selected as the optimizer. And the network is supervised by real-valued ResNet-34 teacher. The weight decay is set following \cite{liu2020reactnet}. The learning rates $\{\eta_1, \eta_2, \eta_3\}$ are set as $\{1e\!\!-\!\!3, 1e\!\!-\!\!4, 1e\!\!-\!\!4\}$ respectively and annealed to $0$ by linear descent.}

We use the Faster-RCNN \cite{ren2016faster} and SSD \cite{liu2016ssd} detection frameworks, which are based on ResNet-18 \cite{he2016deep} and VGG-16 \cite{Simonyan15} backbone, respectively, to train our 
\begin{wrapfigure}{r}{0.5\textwidth}
	\begin{minipage}[t]{0.5\textwidth}
	\subfigure[One-stage]{
		\begin{minipage}[t]{0.5\textwidth}
			\centering
			\includegraphics[width= \linewidth]{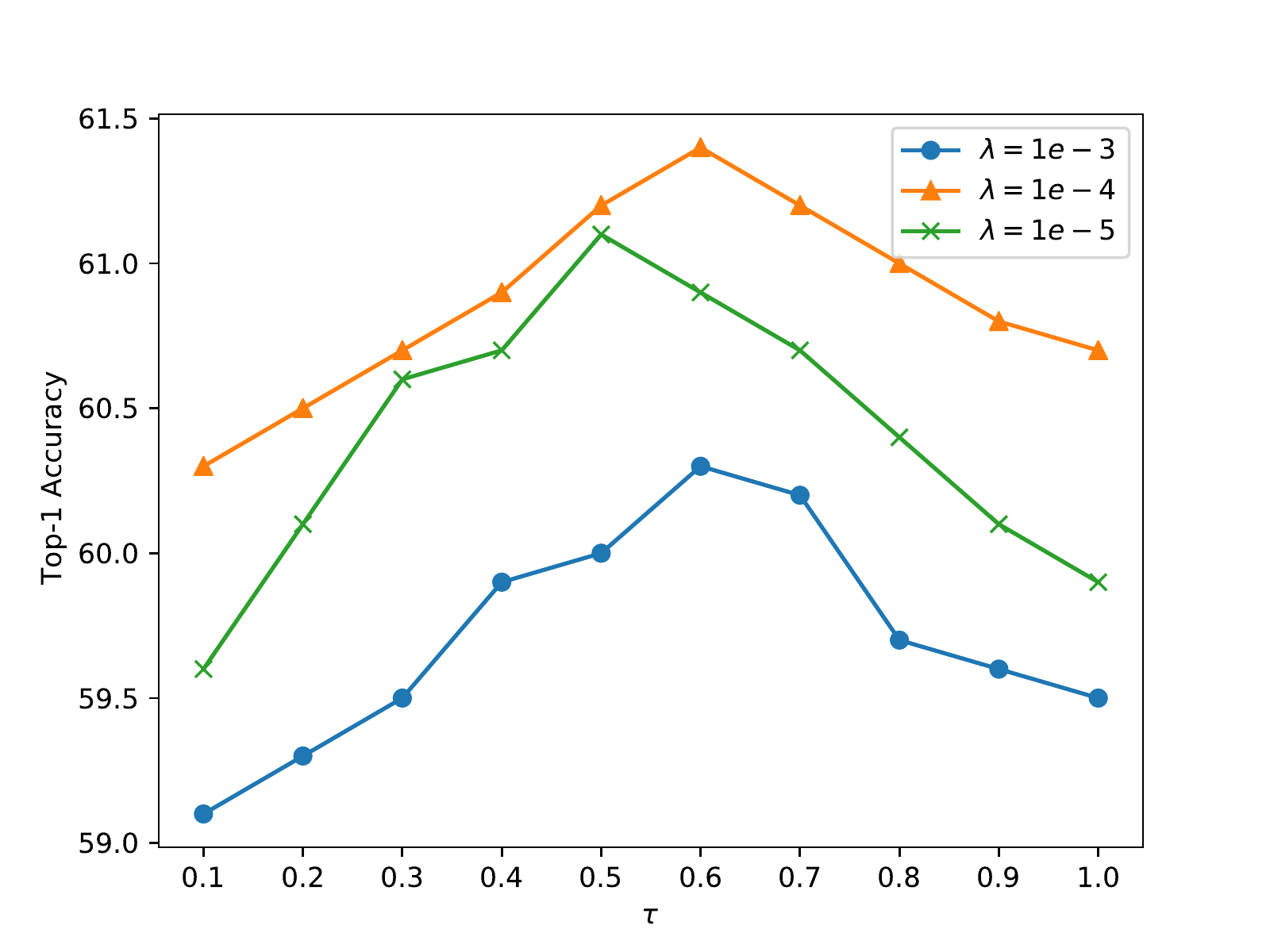}
		\end{minipage}%
	}%
	\subfigure[Two-stage]{
		\begin{minipage}[t]{0.5\textwidth}
			\centering
			\includegraphics[width= \linewidth]{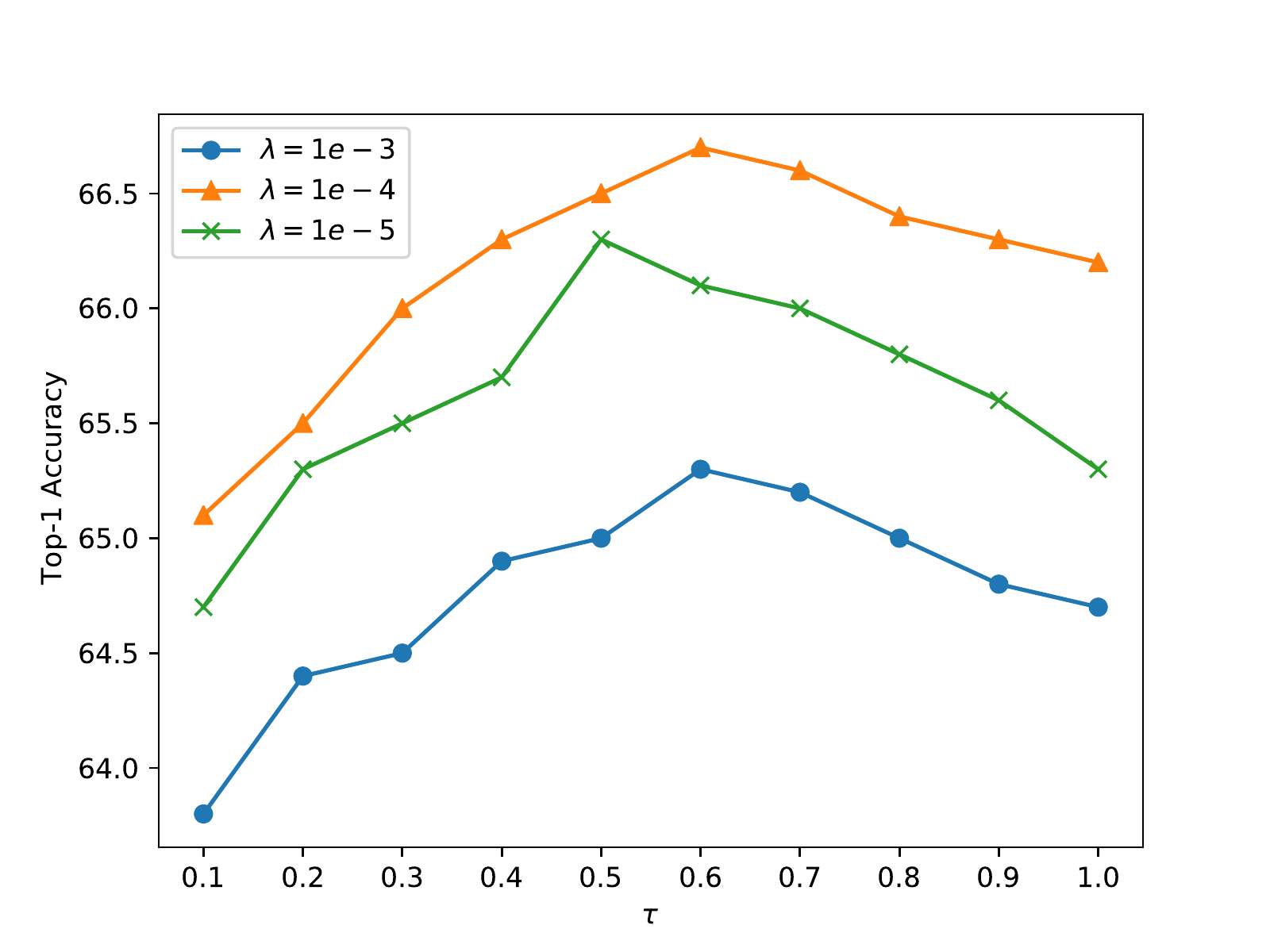}
		\end{minipage}%
	}%
	\caption{Effect of hyper-parameters $\lambda$ and $\tau$ on one-stage and two-stage training using 1-bit ResNet-18.}
	\label{effect}
	\end{minipage}
\end{wrapfigure}
RBONN for the object detection task. We pre-train the backbone for image classification using ImageNet ILSVRC12 \cite{imagenet15} and fine-tune the detector on the dataset for object detection. For SSD and Faster-RCNN, the batch size is set to 16 and 8, respectively, with applying SGD optimizer. Both $\eta_1$ and $\eta_2$ are equal to 0.008. The value of $\eta_3$ is set to 0.001. We use the same structure and training settings as BiDet \cite{wang2020bidet} on the SSD framework.

\subsection{Ablation Study}

\noindent \textbf{Hyper-parameter $\lambda$ and $\tau$.} The most important hyper-parameter of RBONN are $\lambda$ and $\tau$, which control the proportion of $L_R$ and the threshold of backtracking in recurrent bilinear optimization. On ImageNet for 1-bit ResNet-18, the effect of hyper-parameters $\lambda$ and $\tau$ is evaluated under both one-stage and two-stage training. RBONN's performance is demonstrated in Fig. \ref{effect}, where $\lambda$ is varied from $1e{-3}$ to $1e{-5}$ and $\tau$ is varied from 1 to 0.1. As can be observed, with $\lambda$ reducing, performance improves at first before plummeting. When we increase $\tau$ in both implementations, the same trend emerges. As demonstrated in Fig. \ref{effect}, when $\lambda$ is set to $1e{-4}$ and $\tau$ is set to 0.6, 1-bit ResNet-18 generated by our RBONN gets the best performance.  
As a result, we apply this set of hyper-parameters to the remaining experiments in this paper. Note that the recurrent model does not effect when $\tau$ is set as $1$. 

\begin{figure}[t]
	\centering
	\subfigure[Weight distribution]{
		\begin{minipage}[t]{0.5\linewidth}
			\centering
			\includegraphics[width=2.2in]{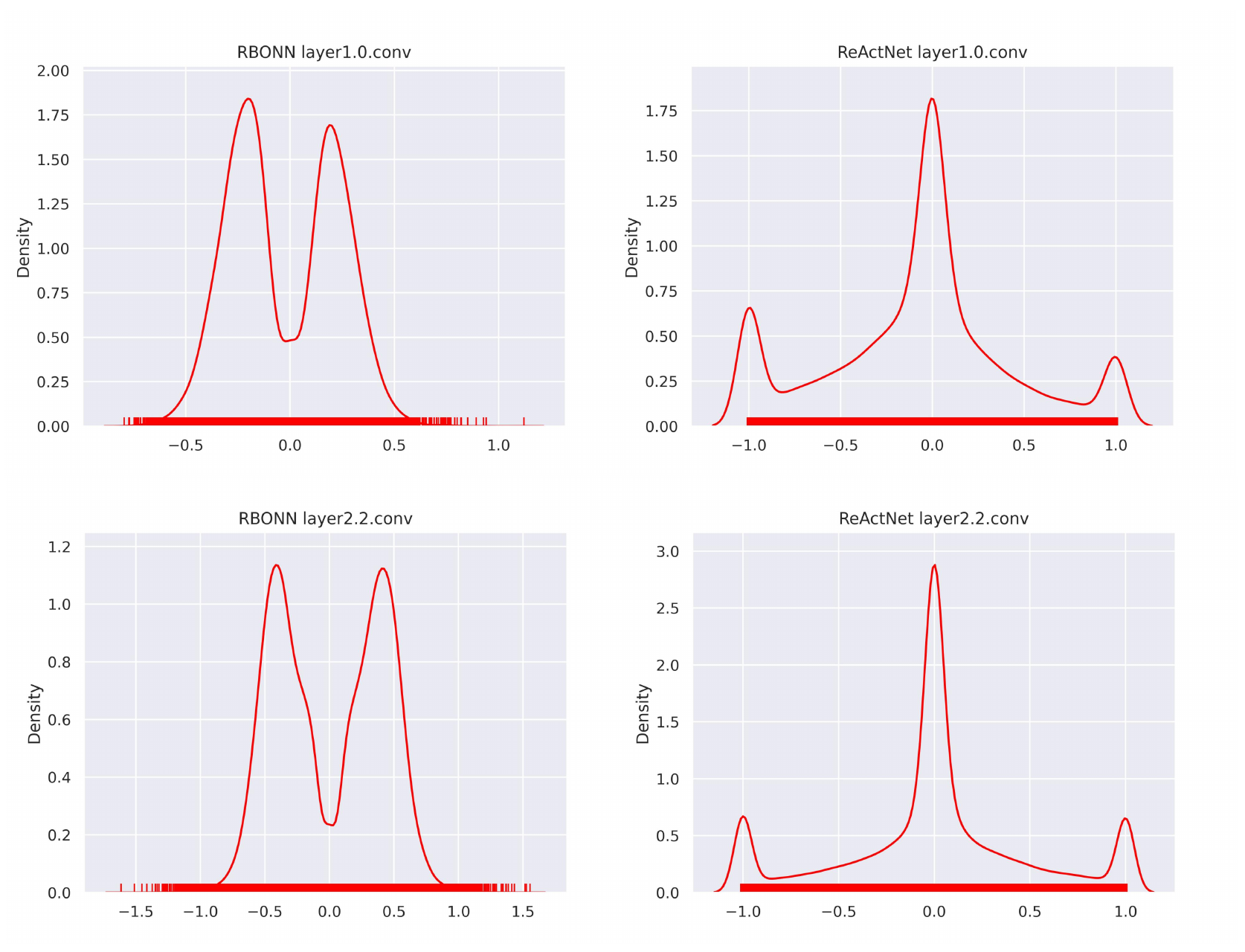}
			%\caption{Diversity analysis on ResNet-20}
		\end{minipage}%
	}%
	\subfigure[Scale matrix distribution]{
		\begin{minipage}[t]{0.5\linewidth}
			\centering
			\includegraphics[width=2.2in]{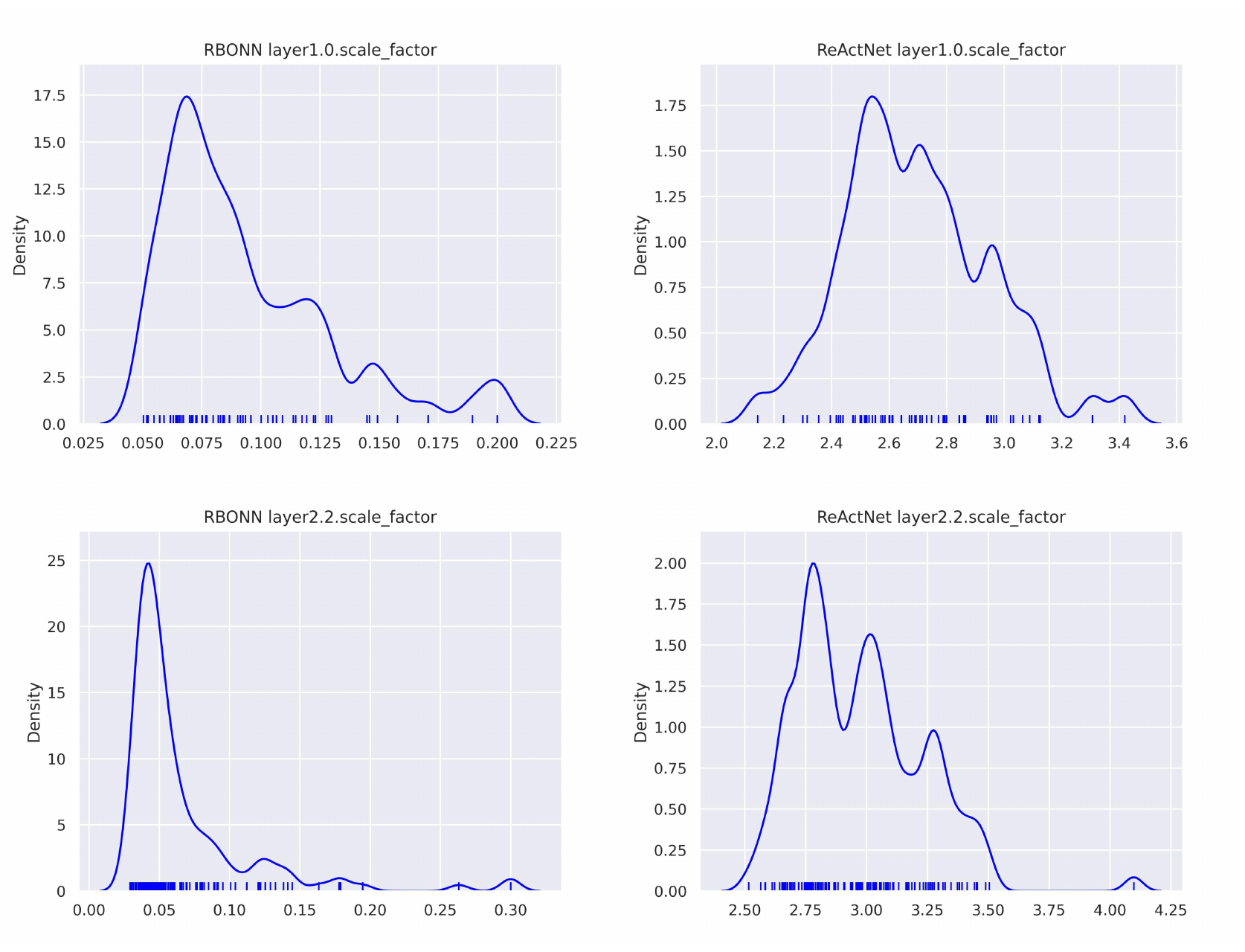}
			%\caption{Diversity analysis on ResNet-32}
		\end{minipage}%
	}%
	\centering
	\caption{Weight (red) and scale matrix (blue) distribution of the RBONN and ReActNet in 1-bit ResNet-18 with two-stage training.}
	%%\vspace{-5mm}
	\label{distribution}
\end{figure}

\noindent \textbf{Weight and {Scale factor} Distribution.} We first analyze the weight distribution of training ReActNet \cite{liu2020reactnet} and RBONN for comparison to analyze the sparsity of ${\bf w}$. For a 1-bit ResNet-18, we analyze the $1$-st and $6$-th 1-bit convolution layer of ResNet-18. \red{The distribution of weights (before binarization) for ReActNet and our RBONN is shown in the left section of Fig. \ref{distribution}. The weight values for ReActNet can be seen to be closely mixed up around the zero centers, and the value magnitude remains sparse.} Thus the binarization results are far less robust to any possible disturbance. In contrast, our RBONN gains weight forming a bi-modal distribution, which achieves its robustness against disturbances. {Moreover, we plot the distribution of non-zero elements in scale matrix $\mathcal{A}$ in the right part of Fig. \ref{distribution}. The scale values of our RBONN is less dense compared with ReActNet. Thus, the result demonstrates that our RBONN prevents $\mathcal{A}$ from going denser and ${\bf w}$ from going sparser, which validates our motivation.}

%%\vspace{-5mm}
\subsection{Image Classification}
\label{sec4.3}
%%\vspace{-2mm}
We first show the experimental results on ImageNet with ResNet-18  \cite{he2016deep} backbone in Tab. \ref{imagenet}. We compare RBONN with BNN \cite{courbariaux2015binaryconnect}, XNOR-Net \cite{rastegari2016xnor}, Bi-Real Net \cite{liu2018bi}, IR-Net \cite{qin2020forward}, BONN\cite{zhao2022towards}, RBCN \cite{liu2019rbcn}, and RBNN \cite{lin2020rotated}. We also report mult-bit DoReFa-Net \cite{zhou2016dorefa}, and TBN \cite{wan2018tbn} for further reference. \red{RBONN outperforms all of the evaluated binary models in both Top-1 and Top-5 accuracy, as shown in Tab. \ref{imagenet}. RBONN achieves 61.4\% and 83.4\% in Top-1 and Top-5 accuracy using ResNet-18, respectively, with 1.8\% and 1.9\% increases over state-of-the-art RBNN.} In this paper, we use memory usage and OPs following \cite{liu2018bi} in comparison to other tasks for further reference. We also analyze the inference speed on hardware in Sec. \ref{deploy}.

Furthermore, we compare with ReActNet \cite{liu2020reactnet} using the same architecture. \red{It uses full-precision parameters, data augmentation, knowledge distillation, and a computationally intensive two-step training setting with 512 epochs in total. We use the same implementation as \cite{liu2020reactnet} to evaluate our RBONN to ReActNet.} As shown in Tab. \ref{react}, our method still achieves an impressive 0.8\% Top-1 accuracy improvement on the same ResNet-18 backbone, which verifies the effectiveness of our method. Also, our method outperforms state-of-the-are ReCU \cite{xu2021recu} by 0.3\% Top-1 accuracy. Moreover, we evaluate the performance of our RBONN on another strong backbone, {\em i.e,}, ReActNet-A. \red{Our strategy improves Top-1 accuracy by 1.2\% on ReActNet-A, which is substantial on the ImageNet dataset classification challenge.}

\red{In a word, when compared to several BNN methods, our RBONN achieves the best performance on the large-scale ImageNet dataset, proving that our method achieves a new state-of-the-art on image classification tasks.}

\begin{table}[t]
\renewcommand\arraystretch{0.86}
\centering
\small
\begin{tabular}{c|c|c|c|c|c}
\hline
Network                     & Method                & W/A                   & \begin{tabular}[c]{@{}c@{}}OPs\\ ($\times10^8$)\end{tabular} & Top-1         & Top-5         \\ \hline
\multirow{11}{*}{ResNet-18} & Real-valued           & 32/32                 & 18.19                                                        & 69.6          & 89.2          \\ \cline{2-6} 
                            & DoReFa-Net            & 1/4                   & 2.44                                                         & 59.2          & 81.5          \\ \cline{2-6} 
                            & TBN                   & 1/2                   & 1.81                                                         & 55.6          & 79.0          \\ \cline{2-6} 
                            & BNN                   & \multirow{8}{*}{1/1} & \multirow{8}{*}{1.63}                                       & 42.2          & 67.1          \\
                            & XNOR-Net              &                       &                                                              & 51.2          & 73.2          \\
                            & Bi-Real Net           &                       &                                                              & 56.4          & 79.5          \\
                            & IR-Net                &                       &                                                              & 58.1          & 80.0          \\
                            & BONN                  &                       &                                                              & 59.3          & 81.6          \\
                            & RBCN                  &                       &                                                              & 59.5          & 81.6          \\
                            & RBNN                  &                       &                                                              & 59.6          & 81.6          \\
                            & \textbf{RBONN} &                       &                                                              & \textbf{61.4} & \textbf{83.5} \\ \hline
\end{tabular}
%%\vspace{2mm}
	\caption{A performance comparison with SOTAs on ImageNet with one-stage training. W/A denotes the bit length of weights and activations. We report the Top-1 (\%) and Top-5 (\%) accuracy performances.}
	\label{imagenet}
	%%\vspace{-6mm}
\end{table}

\begin{table}[t]
\renewcommand\arraystretch{0.86}
\centering
\begin{tabular}{c|c|c|c|c|c}
\hline
Network                    & Method      & W/A                  & OPs($\times10^8$)     & Top-1 & Top-5 \\ \hline
\multirow{4}{*}{ResNet-18} & Real-valued & 32/32                & 18.19                 & 69.6  & 89.2  \\ \cline{2-6} 
                           & ReActNet    & \multirow{3}{*}{1/1} & \multirow{3}{*}{1.63} & 65.9  & -     \\
                           & ReCU    &  &  & 66.4  & 86.5     \\
                           & \textbf{RBONN}       &                      &                       & \textbf{66.7}  & \textbf{87.0}  \\ \hline
\multirow{3}{*}{ReActNet-A}                 & Real-valued & 32/32                & 48.32                 & 72.4  & -     \\ \cline{2-6} 
                           & ReActNet    & \multirow{2}{*}{1/1} & \multirow{2}{*}{0.87} & 69.4  & -     \\
                           & \textbf{RBONN}       &                      &                       & \textbf{70.6}  &    \textbf{89.0}   \\ \hline
\end{tabular}
%%\vspace{2mm}
	\caption{A performance comparison with ReActNet \cite{liu2020reactnet} on ImageNet using two-stage training. W/A denotes the bit length of weights and activations. We report the Top-1 (\%) and Top-5 (\%) accuracy performances.}
	\label{react}
	%%\vspace{-10mm}
\end{table}

%%\vspace{-3mm}
\subsection{Object Detection}
\label{sec4.4}
\noindent\textbf{PASCAL VOC.} 
\red{On the PASCAL VOC datasets, we compare the proposed RBONN against existing state-of-the-art BNNs, such as  XNOR-Net \cite{rastegari2016xnor}, Bi-Real-Net \cite{liu2018bi}, and BiDet \cite{wang2020bidet}, on the same framework for object detection. 
The detection result of multi-bit quantized networks DoReFa-Net \cite{zhou2016dorefa} is also reported. 
In Tab. \ref{VOC}, we show the results for 1-bit Faster-RCNN \cite{ren2016faster} on VOC {\tt test2007} from lines 2 to 7.
With 50.63$\times$ and 19.87$\times$ rate, our RBONN greatly accelerates and compresses the Faster-RCNN with ResNet-18 backbone. We see significant improvements with our RBONN over other methods as compared to 1-bit approaches. With the same memory utilization and FLOPs, our RBONN outperforms XNOR-Net, Bi-Real-Net, and BiDet by 17.0\%, 7.2\%, and 5.9\% mAP, which is substantial on the object detection task.}

\red{For the SSD \cite{liu2016ssd} with VGG-16 backbone, The bottom section in Tab. \ref{VOC} shows that our RBONN can save the computation and storage by 14.76$\times$ and 4.81$\times$, as compared to real-valued alternatives. The difference in performance is rather slight (69.4\% {\em vs.} 74.3\%).} Moreover, compared with other 1-bit SOTAs, our RBONN's performance stands out by a sizeable margin. For example, RBONN surpasses BiDet by 3.4\% with the same structure and compression.

\noindent\textbf{COCO.}
\red{Because of its size and diversity, the COCO dataset presents a greater challenge than PASCAL VOC. On COCO, our RBONN is compared against state-of-the-art BNNs such as XNOR-Net \cite{rastegari2016xnor}, Bi-Real Net \cite{liu2018bi}, and BiDet \cite{wang2020bidet}. We present the performance of the 4-bit DoReFa-Net \cite{zhou2016dorefa} for comparison. Tab. \ref{COCO} does not indicate memory use or FLOPs due to page width constraints. With just different fully-connected layers, the COCO dataset's practical memory utilization and FLOPs are similar to those on VOC.}

\begin{table*}[t]
\renewcommand\arraystretch{0.8}
\centering
\small
\setlength{\tabcolsep}{.5mm}{
\begin{tabular}{c|c|c|c|c|c|c|c}
\hline
Framework                    & \begin{tabular}[c]{@{}c@{}}Input    \\Resolution \end{tabular}                & Backbone                   & Method      & W/A                  & \begin{tabular}[c]{@{}c@{}}Memory Usage\\ (MB)\end{tabular}           & \begin{tabular}[c]{@{}c@{}}OPs\\ ($\times10^9$)\end{tabular}              & mAP     \\ \hline
\multirow{6}{*}{Faster-RCNN} & \multirow{6}{*}{1000$\times$600} & \multirow{6}{*}{ResNet-18} & Real-valued & 32/32                & 47.48                  & 434.39                & 74.6          \\ \cline{4-8} 
                             &                                  &                            & DoReFa-Net  & 4/4                  & 6.73                   & 55.90                 & 71.0          \\ \cline{4-8} 
                             &                                  &                            & XNOR-Net    & \multirow{4}{*}{1/1} & \multirow{4}{*}{2.39}  & \multirow{4}{*}{8.58} & 48.4          \\
                             &                                  &                            & Bi-Real Net &                      &                          &                       & 58.2          \\
                             &                                  &                            & BiDet       &                      &                          &                       & 59.5          \\
                             &                                  &                            & \textbf{RBONN}       &                      &                          &                       & \textbf{65.4} \\ \hline
\multirow{6}{*}{SSD}         & \multirow{6}{*}{300$\times$300}  & \multirow{6}{*}{VGG-16}    & Real-valued & 32/32                & 105.16                 & 31.44                 & 74.3          \\ \cline{4-8} 
                             &                                  &                            & DoReFa-Net  & 4/4                  & 29.58                  & 6.67                  & 69.2          \\ \cline{4-8} 
                             &                                  &                            & XNOR-Net    & \multirow{4}{*}{1/1} & \multirow{4}{*}{21.88} & \multirow{4}{*}{2.13} & 50.2          \\
                             &                                  &                            & Bi-Real Net &                      &                          &                       & 58.2          \\
                             &                                  &                            & BiDet       &                      &                          &                       & 66.0          \\
                             &                                  &                            & \textbf{RBONN}     &                      &                          &                       & \textbf{69.4} \\ \hline
\end{tabular}}
\caption{Comparison of memory usage, OPs, and mAP ($\%$) with state-of-the-art BNNs in SOTA binarized detection frameworks on VOC {\tt test2007}.}
\label{VOC}
%%\vspace{-11mm}
\end{table*}
\begin{table*}[t]
\small
\centering
\setlength{\tabcolsep}{.3mm}{
\begin{tabular}{c|c|c|c|c|c|c|c|c|c}
\hline
Framework                     & \begin{tabular}[c]{@{}c@{}}Input\\Resolution \end{tabular}                 & Backbone                   & Method               & \begin{tabular}[c]{@{}c@{}}mAP\\@{[}.5, .95{]} \end{tabular} & AP$_{50}$       & AP$_{75}$       & AP$_s$       & AP$_m$        & AP$_1$        \\ \hline
\multirow{6}{*}{Faster R-CNN} & \multirow{6}{*}{1000$\times$600} & \multirow{6}{*}{ResNet-18} & Real-valued          & 26.0              & 44.8          & 27.2          & 10.0         & 28.9          & 39.7          \\ \cline{4-10} 
                              &                                  &                            & DeRoFa-Net           & 22.9              & 38.6          & 23.7          & 8.0          & 24.9          & 36.3          \\ \cline{4-10} 
                              &                                  &                            & Xnor-Net             & 10.4              & 21.6          & 8.8           & 2.7          & 11.8          & 15.9          \\
                              &                                  &                            & Bi-Real Net          & 14.4              & 29.0          & 13.4          & 3.7          & 15.4          & 24.1          \\
                              &                                  &                            & BiDet                & 15.7              & 31.0          & 14.4          & 4.9          & 16.7          & 25.4          \\
                              &                                  &                            & \textbf{RBONN} & \textbf{20.6}     & \textbf{37.3} & \textbf{19.9} & \textbf{7.4} & \textbf{21.3} & \textbf{32.8} \\ \hline
\multirow{6}{*}{SSD}          & \multirow{6}{*}{300$\times$300}  & \multirow{6}{*}{VGG-16}    & Real-valued          & 23.2              & 41.2          & 23.4          & 5.3          & 23.2          & 39.6          \\ \cline{4-10} 
                              &                                  &                            & DoReFa-Net           & 19.5              & 35.0          & 19.6          & 5.1          & 20.5          & 32.8          \\ \cline{4-10} 
                                &                                  &                            & XNOR-Net             & 8.1               & 19.5          & 5.6           & 2.6          & 8.3           & 13.3          \\
                              &                                  &                            & Bi-Real Net          & 11.2              & 26.0          & 8.3           & 3.1          & 12.0          & 18.3          \\
                              &                                  &                            & BiDet                & 13.2              & 28.3          & 10.5          & 5.1          & 14.3          & 20.5          \\
                              &                                  &                            & \textbf{RBONN} & \textbf{17.4}     & \textbf{33.2} & \textbf{16.4} & \textbf{5.3} & \textbf{17.1} & \textbf{26.7} \\ \hline
\end{tabular}}
%%\vspace{2mm}
\caption{Comparison of mAP@[.5, .95](\%), AP (\%) with different IoU threshold and AP for objects in various sizes with SOTA 1-bit detectors on COCO {\tt minival}.}
\label{COCO}
\end{table*}

\red{Compared to state-of-the-art XNOR-Net, Bi-Real Net, and BiDet, our method enhances the mAP@[.5,.95] by 10.2\%, 6.2\%, and 4.9\% using the Faster-RCNN framework with the ResNet-18 backbone. 
Moreover, our RBONN clearly outperforms competitors on other APs with various IoU thresholds.
Our RBONN achieves only 2.3\% lower mAP than DeRoFa-Net, a quantized neural network with 4-bit weights and activations.
Our method yields a 1-bit detector with a performance of only 5.4\% mAP lower than the best-performing real-valued counterpart (20.6\% {\em vs.} 26.0\%). Similarly, using the SSD300 framework with the VGG-16 backbone, our method achieves 17.4\% mAP@[.5,.95], outperforming XNOR-Net, Bi-Real Net, and BiDet by 9.3\%, 6.2\%, and 4.2\% mAP, respectively.}

In conclusion, our method outperforms previous BNN algorithms in the AP with various IoU thresholds and AP for objects of various sizes on COCO, demonstrating the method's superiority and applicability in a wide range of applications settings.

\subsection{Deployment Efficiency}
\label{deploy}

\begin{table}[t]
\centering
\begin{tabular}{c|c|c|c|c|c|c}
\hline
Network                     & Method      & W/A   & Size (MB) & Memory Saving             & Latency (ms) & Acceleration              \\ \hline
\multirow{2}{*}{ResNet-18}  & Real-valued & 32/32 & 46.8      & -                         & 1060.2       & -                         \\\cline{2-7}
                            & RBONN       & 1/1   & 4.2       & 11.1$\times$ & 67.1         & 15.8$\times$ \\ \hline
\multirow{2}{*}{SSD-VGG16}  & Real-valued & 32/32 & 105.16& - & 2788.7&-                        \\\cline{2-7}
& RBONN       & 1/1   & 21.88       & 4.8$\times$ &       200.5   & 13.9$\times$ \\ \hline
% \multirow{2}{*}{ReActNet-A} & Real-valued & 32/32 &      117.7     &             -              &     608.1         &         -                  \\
%                             & RBONN       & 1/1   &     7.8      &         15.1\times                  &       144.5       & 4.2\times                          \\ \hline
\end{tabular}
\caption{Comparing RBONN with real-valued models on hardware (single thread).}
\label{hardware}
%%\vspace{-6mm}
\end{table}

\red{We implement the 1-bit models achieved by our RBONN on ODROID C4, which has a 2.016 GHz 64-bit quad-core ARM Cortex-A55. With evaluating its real speed in practice, the efficiency of our RBONN is proved when deployed into real-world mobile devices. 
We leverage the SIMD instruction SSHL on ARM NEON to make the inference framework BOLT \cite{feng2021bolt} compatible with RBONN.
%To make inference framework BOLT \cite{feng2021bolt} compatible with RBONN, we use the SIMD instruction SSHL on ARM NEON. 
We compare RBONN to the real-valued backbones in Tab. \ref{hardware}. We can see that RBONN's inference speed is substantially faster with the highly efficient BOLT framework. For example, the acceleration rate achieves about 15.8$\times$ on ResNet-18, which is slightly lower than the theoretical acceleration rate discussed in Sec. \ref{sec4.3}. Furthermore, RBONN achieves 13.9$\times$ acceleration with SSD. All deployment results are significant for the computer vision on real-world edge devices.}

\section{Conclusion}
This paper proposed a new learning algorithm, termed recurrent bilinear optimization, to efficiently calculate BNNs, which is the first attempt to optimize BNNs from the bilinear perspective. Our method specifically introduces recurrent optimization to sequentially backtrack the sparse real-valued weight filters, which can be sufficiently trained and reach their performance limit based on a controllable learning process. RBONNs show strong generalization to gain impressive performance on both image classification and object detection tasks, demonstrating the superiority of the proposed method over state-of-the-art BNNs.

\paragraph{{\rm {\bf Acknowledgement.}}} This work was supported in part by the National Natural Science Foundation of China under Grant 62076016, 92067204, 62141604 and the Shanghai Committee of Science and Technology under Grant No. 21DZ1100100.

\clearpage
\bibliographystyle{splncs04}
\bibliography{egbib}

\begin{thebibliography}{10}
\providecommand{\url}[1]{\texttt{#1}}
\providecommand{\urlprefix}{URL }
\providecommand{\doi}[1]{https://doi.org/#1}

\bibitem{chen2019metaquant}
Chen, S., Wang, W., Pan, S.J.: Metaquant: Learning to quantize by learning to
  penetrate non-differentiable quantization. Proc. of NeurIPS  \textbf{32},
  3916--3926 (2019)

\bibitem{courbariaux2015binaryconnect}
Courbariaux, M., Bengio, Y., David, J.P.: Binaryconnect: Training deep neural
  networks with binary weights during propagations. In: Proc. of NeurIPS. pp.
  3123--3131 (2015)

\bibitem{courbariaux2016binarized}
Courbariaux, M., Hubara, I., Soudry, D., El-Yaniv, R., Bengio, Y.: Binarized
  neural networks: Training deep neural networks with weights and activations
  constrained to+ 1 or-1. In: Proc. of NeurIPS. pp.~1--9 (2016)

\bibitem{del2011bilinear}
Del~Bue, A., Xavier, J., Agapito, L., Paladini, M.: Bilinear modeling via
  augmented lagrange multipliers (balm). IEEE transactions on pattern analysis
  and machine intelligence  \textbf{34}(8),  1496--1508 (2011)

\bibitem{denil2013predicting}
Denil, M., Shakibi, B., Dinh, L., Ranzato, M., De~Freitas, N.: Predicting
  parameters in deep learning. In: Proc. of NeurIPS. pp. 2148--2156 (2013)

\bibitem{voc2007}
Everingham, M., Van~Gool, L., Williams, C.K., Winn, J., Zisserman, A.: The
  pascal visual object classes (voc) challenge. International Journal of
  Computer Vision  \textbf{88}(2),  303--338 (2010)

\bibitem{feng2021bolt}
Feng, J.: Bolt. \url{https://github.com/huawei-noah/bolt} (2021)

\bibitem{gao2022convmae}
Gao, P., Ma, T., Li, H., Dai, J., Qiao, Y.: Convmae: Masked convolution meets
  masked autoencoders. arXiv preprint arXiv:2205.03892  (2022)

\bibitem{ssegmetation}
Girshick, R., Donahue, J., Darrell, T., Malik, J.: Rich feature hierarchies for
  accurate object detection and semantic segmentation. In: Proc. of CVPR. pp.
  580--587 (2014)

\bibitem{gu2019projection}
Gu, J., Li, C., Zhang, B., Han, J., Cao, X., Liu, J., Doermann, D.: Projection
  convolutional neural networks for 1-bit cnns via discrete back propagation.
  In: Proc. of AAAI. pp. 8344--8351 (2019)

\bibitem{he2015delving}
He, K., Zhang, X., Ren, S., Sun, J.: Delving deep into rectifiers: Surpassing
  human-level performance on imagenet classification. In: Proc. of ICCV. pp.
  1026--1034 (2015)

\bibitem{he2016deep}
He, K., Zhang, X., Ren, S., Sun, J.: Deep residual learning for image
  recognition. In: Proc. of CVPR. pp. 770--778 (2016)

\bibitem{he2018soft}
He, Y., Kang, G., Dong, X., Fu, Y., Yang, Y.: Soft filter pruning for
  accelerating deep convolutional neural networks. In: Proc. of IJCAI. pp.
  2234--2240 (2018)

\bibitem{Huang2017Data}
Huang, Z., Wang, N.: Data-driven sparse structure selection for deep neural
  networks. In: Proc. of ECCV. pp. 304--320 (2018)

\bibitem{lecun1990optimal}
LeCun, Y., Denker, J.S., Solla, S.A.: Optimal brain damage. In: Proc. of
  NeurIPS. pp. 598--605 (1990)

\bibitem{li2016pruning}
Li, H., Kadav, A., Durdanovic, I., Samet, H., Graf, H.P.: Pruning filters for
  efficient convnets. In: Proc. of ICLR. pp. 1--13 (2016)

\bibitem{li2017factorized}
Li, Y., Wang, N., Liu, J., Hou, X.: Factorized bilinear models for image
  recognition. In: Proc. of ICCV. pp. 2079--2087 (2017)

\bibitem{lin2021siman}
Lin, M., Ji, R., Xu, Z., Zhang, B., Chao, F., Xu, M., Lin, C.W., Shao, L.:
  Siman: Sign-to-magnitude network binarization. arXiv preprint
  arXiv:2102.07981  (2021)

\bibitem{lin2020rotated}
Lin, M., Ji, R., Xu, Z., Zhang, B., Wang, Y., Wu, Y., Huang, F., Lin, C.W.:
  Rotated binary neural network. In: Proc. of NeurIPS. pp.~1--9 (2020)

\bibitem{lin2017espace}
Lin, S., Ji, R., Chen, C., Huang, F.: Espace: Accelerating convolutional neural
  networks via eliminating spatial and channel redundancy. In: Proc. of AAAI.
  pp. 1424--1430 (2017)

\bibitem{lin2019towards}
Lin, S., Ji, R., Yan, C., Zhang, B., Cao, L., Ye, Q., Huang, F., Doermann, D.:
  Towards optimal structured cnn pruning via generative adversarial learning.
  In: Proc. of CVPR. pp. 2790--2799 (2019)

\bibitem{coco2014}
Lin, T.Y., Maire, M., Belongie, S., Hays, J., Perona, P., Ramanan, D.,
  Doll{\'a}r, P., Zitnick, C.L.: Microsoft coco: Common objects in context. In:
  Proc. of ECCV (2014)

\bibitem{lin2015bilinear}
Lin, T.Y., RoyChowdhury, A., Maji, S.: Bilinear cnn models for fine-grained
  visual recognition. In: Proc. of ICCV. pp. 1449--1457 (2015)

\bibitem{liu2019rbcn}
Liu, C., Ding, W., Xia, X., Hu, Y., Zhang, B., Liu, J., Zhuang, B., Guo, G.:
  Rbcn: Rectified binary convolutional networks for enhancing the performance
  of 1-bit dcnns. In: Proc. of IJCAI. pp. 854--860 (2019)

\bibitem{liu2019circulant}
Liu, C., Ding, W., Xia, X., Zhang, B., Gu, J., Liu, J., Ji, R., Doermann, D.:
  Circulant binary convolutional networks: Enhancing the performance of 1-bit
  dcnns with circulant back propagation. In: Proc. of CVPR. pp. 2691--2699
  (2019)

\bibitem{liu2018darts}
Liu, H., Simonyan, K., Yang, Y.: Darts: Differentiable architecture search. In:
  Proc. of ICLR (2019)

\bibitem{liu2016ssd}
Liu, W., Anguelov, D., Erhan, D., Szegedy, C., Reed, S., Fu, C.Y., Berg, A.C.:
  Ssd: Single shot multibox detector. In: Proc. of ECCV. pp. 21--37 (2016)

\bibitem{liu2020reactnet}
Liu, Z., Shen, Z., Savvides, M., Cheng, K.T.: Reactnet: Towards precise binary
  neural network with generalized activation functions. In: Proc. of ECCV. pp.
  143--159 (2020)

\bibitem{liu2018bi}
Liu, Z., Wu, B., Luo, W., Yang, X., Liu, W., Cheng, K.T.: Bi-real net:
  Enhancing the performance of 1-bit cnns with improved representational
  capability and advanced training algorithm. In: Proc. of ECCV (2018)

\bibitem{Liu2017learning}
Liu, Z., Li, J., Shen, Z., Huang, G., Yan, S., Zhang, C.: Learning efficient
  convolutional networks through network slimming. In: Proc. of ICCV. pp.
  2736--2744 (2017)

\bibitem{paszke2017automatic}
Paszke, A., Gross, S., Chintala, S., Chanan, G., Yang, E., DeVito, Z., Lin, Z.,
  Desmaison, A., Antiga, L., Lerer, A.: Automatic differentiation in pytorch.
  In: NeurIPS Workshops (2017)

\bibitem{petersen2008matrix}
Petersen, K., Pedersen, M., et~al.: The matrix cookbook. Technical University
  of Denmark  \textbf{15} (2008)

\bibitem{qin2020forward}
Qin, H., Gong, R., Liu, X., Shen, M., Wei, Z., Yu, F., Song, J.: Forward and
  backward information retention for accurate binary neural networks. In: Proc.
  of CVPR. pp. 2250--2259 (2020)

\bibitem{rastegari2016xnor}
Rastegari, M., Ordonez, V., Redmon, J., Farhadi, A.: Xnor-net: Imagenet
  classification using binary convolutional neural networks. In: Proc. of ECCV.
  pp. 525--542 (2016)

\bibitem{ren2016faster}
Ren, S., He, K., Girshick, R., Sun, J.: Faster r-cnn: Towards real-time object
  detection with region proposal networks. IEEE Transactions on Pattern
  Analysis and Machine Intelligence  \textbf{39}(6),  1137--1149 (2016)

\bibitem{romero2014fitnets}
Romero, A., Ballas, N., Kahou, S.E., Chassang, A., Gatta, C., Bengio, Y.:
  Fitnets: Hints for thin deep nets. In: Proc. of ICLR. pp. 1--13 (2015)

\bibitem{imagenet15}
Russakovsky, O., Deng, J., Su, H., Krause, J., Satheesh, S., Ma, S., Huang, Z.,
  Karpathy, A., Khosla, A., Bernstein, M., Berg, A.C., Fei-Fei, L.: Imagenet
  large scale visual recognition challenge. International Journal of Computer
  Vision  \textbf{115}(3),  211--252 (2015)

\bibitem{Simonyan15}
Simonyan, K., Zisserman, A.: Very deep convolutional networks for large-scale
  image recognition. In: Proc. of ICLR. pp. 1--13 (2015)

\bibitem{suh2018part}
Suh, Y., Wang, J., Tang, S., Mei, T., Mu~Lee, K.: Part-aligned bilinear
  representations for person re-identification. In: Proc. of ECCV. pp.
  1449--1457 (2018)

\bibitem{wan2018tbn}
Wan, D., Shen, F., Liu, L., Zhu, F., Qin, J., Shao, L., Tao~Shen, H.: Tbn:
  Convolutional neural network with ternary inputs and binary weights. In:
  Proc. of ECCV. pp. 315--332 (2018)

\bibitem{wang2020bidet}
Wang, Z., Wu, Z., Lu, J., Zhou, J.: Bidet: An efficient binarized object
  detector. In: Proc. of CVPR. pp. 2049--2058 (2020)

\bibitem{xu2021poem}
Xu, S., Li, Y., Zhao, J., Zhang, B., Guo, G.: Poem: 1-bit point-wise operations
  based on expectation-maximization for efficient point cloud processing. In:
  Proc. of BMVC. pp. 1--10 (2021)

\bibitem{xu2021layer}
Xu, S., Zhao, J., Lu, J., Zhang, B., Han, S., Doermann, D.: Layer-wise
  searching for 1-bit detectors. In: Proc. of CVPR. pp. 5682--5691 (2021)

\bibitem{xu2021recu}
Xu, Z., Lin, M., Liu, J., Chen, J., Shao, L., Gao, Y., Tian, Y., Ji, R.: Recu:
  Reviving the dead weights in binary neural networks. In: Proc. of ICCV. pp.
  5198--5208 (2021)

\bibitem{yang2020searching}
Yang, Z., Wang, Y., Han, K., Xu, C., Xu, C., Tao, D., Xu, C.: Searching for
  low-bit weights in quantized neural networks. In: Proc. of NeurIPS. pp. 1--11
  (2020)

\bibitem{yu2017multi}
Yu, Z., Yu, J., Fan, J., Tao, D.: Multi-modal factorized bilinear pooling with
  co-attention learning for visual question answering. In: Proc. of ICCV. pp.
  1821--1830 (2017)

\bibitem{zhao2022towards}
Zhao, J., Xu, S., Zhang, B., Gu, J., Doermann, D., Guo, G.: Towards compact
  1-bit cnns via bayesian learning. International Journal of Computer Vision
  \textbf{130}(2),  201--225 (2022)

\bibitem{zhou2016dorefa}
Zhou, S., Wu, Y., Ni, Z., Zhou, X., Wen, H., Zou, Y.: Dorefa-net: Training low
  bitwidth convolutional neural networks with low bitwidth gradients. arXiv
  preprint arXiv:1606.06160  (2016)

\end{thebibliography}
\end{document}